\documentstyle[12pt]{article}
\begin{document}
\thispagestyle{empty}

\title{Phonetic Ambiguity : Approaches, Touchstones, Pitfalls, and New
Directions}
\author{{\em Patrick Juola} \\
Department of Experimental Psychology \\ Oxford University \\
South Parks Road \\ Oxford, UK  OX1 3UD \\ {\tt patrick.juola@psy.ox.ac.uk } }

\date{}

\addtolength{\topmargin}{-0.75in}
\addtolength{\textheight}{1.5in}

\pagestyle{empty}

\maketitle

\begin{abstract}
Phonetic ambiguity and confusibility are bugbears for
any form of bottom-up or data-driven approach to language processing.
The question of when an input is ``close enough'' to a target word
pervades the entire problem spaces of speech recognition, synthesis,
language acquisition, speech compression, and language representation, 
but
the variety of representations that have been applied are demonstrably
inadequate to at least some aspects of the problem.
This paper reviews this inadequacy by
examining several touchstone models in phonetic
ambiguity and relating them to the problems they were designed to solve. 
An good solution would be, among other things, efficient, accurate,
precise, and universally applicable to representation of words, ideally
usable as a ``phonetic distance'' metric for direct measurement of
the ``distance'' between word or utterance pairs.
None of the proposed models can provide a
complete solution to the problem;
in general, there is no algorithmic
theory of phonetic distance.   It is unclear
whether this is a weakness of our representational technology or a more
fundamental difficulty with the problem statement.
In any case, these results show that the representations can be as
crucial as the system architecture, and that as much or more creativity
is required to properly represent language as to process it.
\end{abstract}

\section{Motivations}
When is a sound one word and not another?  This is, in many regards,
{\em the} fundamental question for spoken language processing.  At
a minimum, any system must be capable of converting the infinite
variety of human speech into a distinct set of words, regardless
of the subsequent processing to be performed.  Somehow, a system
must distill the
relevant and important information from the highly noisy and
ambiguous sound signals.  And, although systems will differ slightly
on what are considered to be ``relevant,'' certain characteristics
appear to be universal, regardless of the system/model---and thus
it makes sense to talk about an ``ideal'' representation, suited
to any purpose.

What properties would an ideal sound and language representation scheme
have? As a first pass it should be capable of accurately
representing the important contrasts.  It would be
broadly useful
to the exact extent that it can represent a wide and hopefully universal
range of words.   An ideal representation should be useful for as wide
a range of {\em problems} as possible, including the modelling of
human language processing.
And for many applications, it should
be reversible so that representations can
be reconverted in words at the end of some form of processing.

Efficiency of storage and manipulation is another
desideratum, as language applications often require that
thousands of words and millions of sentences be stored and analyzed.
It would be useful if the representation could be related to
psycholinguistic perception and/or production theories.
Similarly, the more
accurately the representation can capture cognitive facts, the fewer
representational
artifacts it will introduce into (e.g.) modelling work.
Another important characteristic of an ideal system is what it does
{\em not} represent.  For most applications aspects of
speech such as sex, health, and accent of the speaker should be
ignored, or in other words, the representation should be speaker-independent.
For engineering reasons, the representation should be modular and
reusable, so that the same wheel need not be continually reinvented.
Finally, it should also be symbolically decomposable,
so that, a researcher can knowledgeably manipulate the effects of
specific phenomena such as stress, prosody, or accent.

A rather more controversial claim is that the the representation
should include a meaningful distance measurement.
For most applications, it would be convenient if
one sound pattern could be said to be ``closer'' to a desired
word than another -- e.g. providing guidance in a language learning
experiment.  A distance metric could aid case-based reasoning and/or
error recovery by providing a measure 
of confidence, confusibility, or likelihood.
However, to conjecture
that phonetics can be approximated by a metric space is a very strong
mathematical statement; although such would be
desirable, much further work would be required to firmly establish that
claim.  However, this property is important enough for engineering reasons
that it seem reasonable to include this as a significant factor in
evaluating representation systems.

\section{Sound signals}

The simplest and easiest way of representing sound is to
use the acoustic signal itself. 
No technique can be more faithful of reproduction or easily
reversible.  Cognitive plausibility is immediate and obvious, and
application is clearly universal.  Applications
in speech processing that use the speech signal itself without
a separate unit performing preprocessing are too many, common, and
varied to require individual mention; any recent COLING, ICSLP, or ACL
proceedings will list several applications.

Unfortunately, this is one of the least efficient representations
available, in terms of
both storage and processing.
It is also legendary for being speaker-dependent, enough so
to be used as the key to several security systems (e.g.
PGPfone\cite{pgpfone}).  A more serious objection to its use is
that the usual purpose for language representations is
to simplify the task of processing by eliminating extraneous, useless,
or noisy information.  Any two applications are likely to jointly consider at
least some of the information, such as speaker-dependence, to be
extraneous, and thus the sound signal itself is best treated
as a base case, or as a reserve representation when no other one
presents itself.

\section{Russell/SOUNDEX}

One of the earliest attempts to model phonetic ambiguity derives from
the representation of names.  For example, an important
letter addressed to the ``Brown'' family may actually be intended for
the Braun family.
The Russell/SOUNDEX
encoding (described in \cite{soundex}) provides a method of encoding
names to reduce or eliminate misclassification errors caused
by similar sounding names or transcription errors.

SOUNDEX represents words (usually names) as four character strings.  The
first character of the name becomes the first character of the code
string.  Every subsequent character in the name is looked up according
to the scheme presented as
Figure~\ref{figure:soundex} and 
encoded as one of the digits 1-6.  The first three (coded) characters
are appended to the initial letter and 
are used as the word coding---if there are insufficient characters remaining in the
name, the coding is padded with zeros.
Two adjacent identically coded letters
(for example, the double ll in ``Miller'') are treated as a single
letter.

\begin{figure}
\centering
\begin{tabular}{ll}
     1 = B,P,F,V & 4 = L \\
     2 = C,S,G,J,K,Q,X,Z & 5 = M,N \\
     3 = D,T & 6 = R \\ \\
\multicolumn{2}{c}{All other letters (A,E,I,O,U,Y,W,H) ignored}
\end{tabular}
\caption{\label{figure:soundex}Russell/SOUNDEX coding guide}
\end{figure}

For example, the name ``Juola'' would be coded as J400.  The initial
J is transcribed unchanged, the U and the O are ignored, the L is
coded as a 4, the A is ignored, and the code is padded with zeros to
a full four characters.  Similarly,
``Krumplestater'' and ``Kruempelstaedter'' would both be coded
as K651.

Within the limited domain of filing names, SOUNDEX works extremely well.
SOUNDEX coding is simple to perform by hand, efficient to implement
on a computer, and robust to most transcription and
misspelling errors.
Furthermore, it is robust to most accent or pronunciation variations,
meaning that it will integrate well into a text-based system with
voice input, such
as an (automated) airline reservation system.
Names of arbitrary length are conveniently compressed to a uniform
size and format.

However, the number of false positives, names that are incorrectly
grouped together, is much higher.  For example, B560 codes the ``Bonner''
variants as well as ``Baymore.''  V525 is ``Van Hoesen''
as well as ``Vincenzo.''
There being fewer than 9000 classes in total, 
a large number of false-positive errors is almost certain to arise.
Furthermore, names in certain categories can cause an
unacceptable amount of clustering, and it has proved necessary for
some applications to significantly modify the SOUNDEX coding (e.g.
the Daitch-Motokoff variant for Jewish names\footnote{see
{\em http://http://www.genealogysf.com/glenda.html}}).
The application
of the SOUNDEX scheme is thus highly use specific and marginally
language specific as well.

The greatest problem with the universal application of a SOUNDEX-like
encoding to language problems is 
the absence of a distance measure.
As with a hash table, there's no sense in which numbers/names
in one category can be meaningfully stated to be close or distant from
another category -- or even in which categories can meaningfully
be considered to be equivalence classes.  The coding scheme is
non-invertible; there is no way to
reconstruct a name from its SOUNDEX category. 
Finally, although the system is relatively robust to spelling
variations, there are certain
categories of errors to which it is extremely sensitive.  For example,
changing the initial consonant of a name (``Cramer'' to ``Kramer'') or
insertion or deletion of a letter (``Boughman'' to ``Bowman'') can
change the coding of the name, and there's no practical method of
identifying a set of ``neighboring categories'' into which a (mis)coding
might fall.

\section{Templates and PGPfone}
Another simple method of performing word to word comparisons is
to simply divide word (pairs) into  elements and perform comparisons
on those elements.
Phonetic theory (e.g. \cite{ladefoged})
provides  support for the notion of analyzing words as a
temporal sequence
of phonemes which can be individually compared.
For instance the first elements (phonemes)
can be compared, then the second, \&c.,
and a total distance calculated as a function of the element distances.

This technique has been in common use in, e.g., neural
net\cite{cottrell-plunkett} research for a number of years.  A
detailed example
of this technique in use can be found in the PGPfone
alphabet\cite{juola-nemlap2,pgpfone}.   This alphabet is
the result of a computer search for ``phonetically distinct'' words
using an elaborate feature-based metric.  Phonetic
features, as described in \cite{ladefoged}, are compared and the
results of the comparison are weighted in approximate accord with
psycholinguistic results on perceptual salience, as typified
by \cite{miller-nicely}.  Figure~\ref{table:features} describes this
weighting in detail.
\begin{figure}
\centering
\begin{tabular}{lll}
Feature name & Sample & Weighting \\ \\
Place of articulation & /d/ vs /g/ & 7 \\
Manner of articulation & /l/ vs /t/ & 6 \\
Height of articulation & /i/ vs /$\epsilon$/ & 5 \\
Voicing & /z/ vs /s/ & 4 \\
Syllabic & vowels vs. cons. & 1 \\
Nasal & /n/ vs /d/ & 1 \\
Lateral & /l/ vs /r/ & 1 \\
Roundedness & (various) & 1 \\
Sibilant & /s/ vs /f/ & not used \\
\end{tabular}
\caption{\label{table:features}Phoneme coding for PGPfone alphabet}
\end{figure}
Suprasegmental features such as stress pattern and the additional
salience of the onset consonants are incorporated by a second
level of reweighting.  The various weightings can, in theory, be
arbitrarily refined to match available psycholinguistic data.

Because the PGPfone metric was specifically designed as a distance
metric, it can be used directly to measure closeness of fit or
accuracy of a word coding.
It is efficient to code and decode words,
and the accuracy is as good
as the psycholinguistic data it incorporates.  Similarly, it merges
easily with existing fields of study.  It is as language, dialect, and
speaker-independent as the underlying phonological representation.

In some regards, this very accuracy can be a weakness, as it can
put more stress on the psycholinguistic data than the data can
support.  For example, \cite{ladefoged} and similar feature sets
usually describe phonemes in terms of productive differences; subtle
differences such as tongue placement and place of articulation are
described with much more accuracy and detail than binary features such
as voicing.  However, studies such as \cite{miller-nicely} indicate that,
for example, voicing is usually more salient than place of
articulation, and thus the word ``but'' is more likely to be
misheard as ``gut'' than as ``putt'' under nearly all noise conditions.
However, data have only been gathered and collated for
a fraction of the relevant comparison conditions; in the case of
\cite{miller-nicely}, for instance, only consonant comparisons
were done.

A more serious weakness of the PGPfone scheme lies in the
limited domain of comparison.  Comparison on a phoneme by
phoneme basis requires that individual phonemes to compare be aligned
meaningfully.  For words with an identical number of phonemes in a
sufficiently similar pattern, this is nonproblematic.  However, in the
general case, there's no easy and efficient algorithmic method to
compare one consonant against a cluster, or a two syllable word against
a five syllable one.  Which, for instance, should be measured as
closer to the word "bet", the word "bets,", "best", or "Bess"?
PGPfone, like most other applications of this technique,
avoids this problem by carefully restricting
the domain of comparisons (for example, only words with an identical
number of syllables can be meaningfully compared).  This restriction
can, in the domain of PGPfone, be turned into an advantage by carefully
phrasing the restrictions to limit its use to words advantageous for other
purposes,
but in the general case, this limitation can only be overcome
by lots of additional computation to determine the appropriate alignment,
resulting in computational inefficiency.  Again one can observe that
this line of approach has serious flaws in the way of developing an ideal
universal phonetic representation system.

\section{Autosegmental representations}
The assumption of the previous section, that words are a linear
sequence of elements, may produce raised eyebrows for some modern
phonologists.
Autosegmental phonology \cite{goldsmith})
describes words in terms of 
multiple `tiers' of different parallel phenomena, and the horizontal
slicing into tiers is primary to the vertical time-slicing of the
sound sequence itself.  Words can thus be identified, classified, and compared
at several levels.  This can make the classification
more robust to minor changes at lower levels (such as insertion
of a single consonant).

Bird and Ellison\cite{bird-ellison} describe a method for algorithmically
representing words in this fashion that can also be applied
to determining interword compatibility at the various levels.  They are
content to restrict themselves to merely demonstrating the algorithm in
use and to use it to develop and demonstrate a few phonological rules.
The rules they develop are of the same sort used by several other
researchers (e.g. \cite{cottrell-plunkett}) in 
models of the production of past tenses.  This is another classic
touchstone problem for the testing of cognitive theories of language.

Again, these models merge well with existing phonological theory and
can be used to represent any word in any language.  The
representations are accurate and algorithmic, with good modularity and
speaker-independence.  Unfortunately, the encodings for these
representations (as finite-state automata) yield tremendously
inefficient algorithms, even for determining whether or not two
representations describe the same words.  

In \cite{bird-ellison}, for instance, autosegmental tiers within
a word are associated by describing each tier with a separate
automaton, calculating ``pinnings'' between the tiers as state
numberings, and then determining the final representation as
a larger automaton that accepts all and only the intersection
of the languages accepted by the various tiers.  This method can
clearly be generalized to the comparison of multiple words by
determining appropriate pinnings between the representations of
two words and determining whether the intersection of all tiers,
of both words, is non-empty.  However, the calculation of intersection
of such automata is typically polynomial in the number of automata,
even ignoring the difficulties in determining appropriate pinnings
(as alluded to in the previous section).  This formalism is
thus difficult to efficiently implement on a large-scale, as would
be necessary for case-based reasoning, \&c.

\section{Discussion and conclusions}

All representations and codings discussed (with the exception of the
null encoding) manage to produce contrastive representations
in a speaker-independent fashion.  This should not
be particularly surprising, as any representation that could not
distinguish relevant sounds simply wouldn't work.  On the other
hand, it appears difficult if not impossible to simultaneously
satisfy all three of the desiderata of
algorithmic efficiency, generality of representation, and accuracy of
representation. 
In particular, the SOUNDEX coding uses a very coarse and inaccurate
representation, with no useful topological properties, but can be
efficiently and effectively implemented on nearly any set of words
in English.  The PGPfone templates provide an efficient and detailed
distance metric between any two words in an extremely restricted class
selected for other engineering reasons.  Autosegmental encoding, as
developed by \cite{bird-ellison}, is in theory extremely accurate and
can capture any phonological variation of significance, but is
algorithmically so inefficient as to have prevented it from being used
in any major projects.  None of the three approaches is capable of
quickly determining an ambiguity or confusion measure between a pair
of random English words.

From an engineering perspective, this is disappointing but perhaps
unsurprising; the idea of ``pick two and call me back'' is a common
joke.  From a scientific point of view, the implications are more
interesting.  If the notion of an ideal phonological distance
metric is well-founded and achievable, then the obvious question to
be addressed is how to find it and what sort of data are necessary
to gather.  If, on the other hand, it is not possible to develop
an ideal metric, what is the human solution?  For example, if
human language processing is not done speaker-independently, then
how does the speaker influence the processing and representation
of sounds in the human brain?  What are the psychological and
linguistic implications of these representations?  If there is
no possible metrization of perceptual errors, then where do the
different confusibilities arise?  What sort of phenomena can
be expressed in a psycholinguistically real fashion?

In the shorter term, these results indicate once again the importance
of task and problem analysis for language problems.
Many  researchers are
content, in the absence of a better metric, to treat the results of their
neural networks or cluster analyses of phonetic templates as being
representative of cognitive processes within the human brain.  A better
approach might be to treat the representation process itself with as
much caution and creativity as the system architecture, rather than trusting
in crude representational simplifications and hoping that these
simplifications are nondestructive.  A careful task analysis can, one
hopes, demonstrate not only what aspects of ambiguity processing are
relevant for an engineering solution to a language problem, but also
the ways in which solutions to different touchstone problems, even
apparently arbitrary ones such as PGPfone or SOUNDEX,  can be
fitted together as a scaffold for the larger psycholinguistic questions.

\end{document}